\documentclass[runningheads]{llncs}

\usepackage{eccv}

\usepackage{eccvabbrv}

\usepackage{graphicx}
\usepackage{booktabs}
\usepackage{multirow}
\usepackage[inline]{enumitem}
\usepackage{xcolor}
\usepackage{array}
\usepackage{graphicx,caption}
\newcolumntype{M}[1]{>{\centering\arraybackslash}m{#1}}
\def\BibTeX{{\rm B\kern-.05em{\sc i\kern-.025em b}\kern-.08em
    T\kern-.1667em\lower.7ex\hbox{E}\kern-.125emX}}

\usepackage[accsupp]{axessibility}  % Improves PDF readability for those with disabilities.

\usepackage{hyperref}

\usepackage{orcidlink}

\begin{document}

\title{UniCon-Former: Unified Convolution Transformer is All You Need for Hand Gesture Recognition} 

\titlerunning{UniCon-Former: Unified Convolution Transformer}

\author{Mallika Garg\inst{1}\orcidlink{0000-0002-6056-6490} \and
Debashis Ghosh\inst{2}\orcidlink{0000-0001-5672-7645} \and
Pyari Mohan Pradhan\inst{2}\orcidlink{0000-0002-6070-5577
}}

\authorrunning{M.~Garg et al.}

\institute{Indian Institute of Technology Kharagpur, INDIA \and
Indian Institute of Technology Roorkee, INDIA\\
\email{mallika@ec.iitr.ac.in,  debashis.ghosh@ece.iitr.ac.in,
pyarimohan.pradhan@gmail.com}\\
% \url{http://www.springer.com/gp/computer-science/lncs}
}

\maketitle

\begin{abstract}
Convolutional Neural Networks (CNNs) capture local features efficiently but struggle with global context due to their limited receptive field. On the other hand, transformers effectively capture global dependencies through self-attention but suffer from high redundancy and computational costs. Thus, to leverage the advantages of both CNNs and transformers, we propose a unified model (UniCon-Former) that aims to provide robust and efficient performance on dynamic hand gesture recognition. The unified approach helps the model to learn both local and global features. At the beginning of each transformer stage, the convolution projections help in decreasing the dimension of the input vectors of the transformer block. This creates a pyramidal structure at each transformer stage. These features enable the UniCon-Former to reduce resource usage than vanilla transformers, making it flexible for learning multi-scale and high-resolution features, which is required in hand gesture recognition. We have performed experiments with NVGesture and Briareo datasets and achieved state-of-the-art results with fewer parameters and MACs.
\end{abstract}

\section{Introduction}
\label{sec:intro}
Hand gesture recognition has captured researchers' attention as it can be used in many applications like Human-computer interaction, Gaming, 
Virtual and augmented reality~\cite{ohkawa2023assemblyhands}, 
Sign language recognition~\cite{kumar2017coupled},
% Automotive~\cite{cisotto2020feature},
Healthcare~\cite{zhao2018mobigesture}.
% Robotics~\cite{van2021gesture}, 
% Autonomous vehicles~\cite{garg2021deep}. 
Gestures can be static or dynamic. Static gestures involve holding a specific hand, and dynamic gestures involve continuous movement of body parts, particularly the hands and arms, to convey meaning or information.
% These gestures typically involve movements and hand or body position changes over time. In this paper, we focus our implementation on dynamic gestures which are sequential data.

% Initially, traditional methods were used for hand recognition based on different classifiers such as SVM, Naive-Bayes classifier, etc using hand-crafted features like joint angles~\cite{avola2018exploiting}, color-coded joint velocity feature map~\cite{kumar2018three}, template matching, edge detection etc.
% Recent trends in hand gesture recognition include multimodal approaches combining RGB and depth information, real-time performance on embedded devices, and robustness to varying environmental conditions and hand poses. However, 
With the advancements in vision-based pattern recognition technology, researchers have increasingly shifted towards using self-learned features extracted by deep learning models
% ~\cite{9691523}
. Today, more research is being done to design an efficient transformer model with comparable or better performance. To reduce the computational complexity, self-attention in a transformer is replaced by a generalized concept of token mixer
% ~\cite{yu2022metaformer} 
which says that transformers need a mixer to facilitate communication and interaction between tokens (or patches) in the input sequence. These mixers enable the exchange of information among tokens, allowing the model to capture dependencies and relationships across the input sequence more effectively. Also, despite the revolution in transformer-based methods, their utility in gesture recognition is very limited. In~\cite{d2020transformer}, the vanilla transformer recognizes dynamic hand gestures. This model learns long-range dependencies and global context that help the model learn highly flexible hand shapes and adapt to various hand sizes without requiring significant architectural changes. But the self-attention mechanism in transformer can lead to high redundancy because it performs blind similarity comparisons among all tokens, which can be computationally expensive and inefficient.

To address this limitation, both CNN and transformer are used in combination to integrate their respective strengths into a unified framework. Although much work has been done on models that aggregate convolution with the transformer model in the field of Visual Recognition~\cite{li2023uniformer},  Medical Image
Segmentation~\cite{lin2023convformer}, Scene Text Understanding~\cite{deshmukh2024textual}.  But still this combination has not been explored in the literature for gesture recognition tasks. We combine the strengths of convolution and the transformer.  At each stage of the transformer, the input to the transformer progressively decreases with the use of convolution before feeding the input to the transformer.  This helps to create a pyramid structure of the transformer stage and learn multiscale features by progressively shrinking the attention dimension at each stage.

Thus, the proposed UniCon-Former has the following   major contributions:
\begin{enumerate}
    \item A novel \textbf{U}\textbf{ni}fied \textbf{Con}volution Trans\textbf{former}, \textbf{UniCon-Former} network  for dynamic hand gesture recognition.
    
    \item We combine convolution with the transformer model to leverage the strengths of CNNs, such as capturing local features while also incorporating the benefits of transformers, including capturing global context information.
    
    \item The pyramid hierarchy of features learned at different stages of the transformer helps the model to learn multiscale features which play a significant role in gesture recognition, as hand shape and size can vary, which also adds to the decrease of the computation cost.
    
    \item The effectiveness of the proposed framework is validated using two publicly available datasets: the NVidia Dynamic Hand Gesture and the Briareo dataset. Our model achieves state-of-the-art results when compared to existing methods with fewer parameters and less complexity.
\end{enumerate}

\begin{figure*}[t]			
\centerline{\includegraphics[scale=0.4]{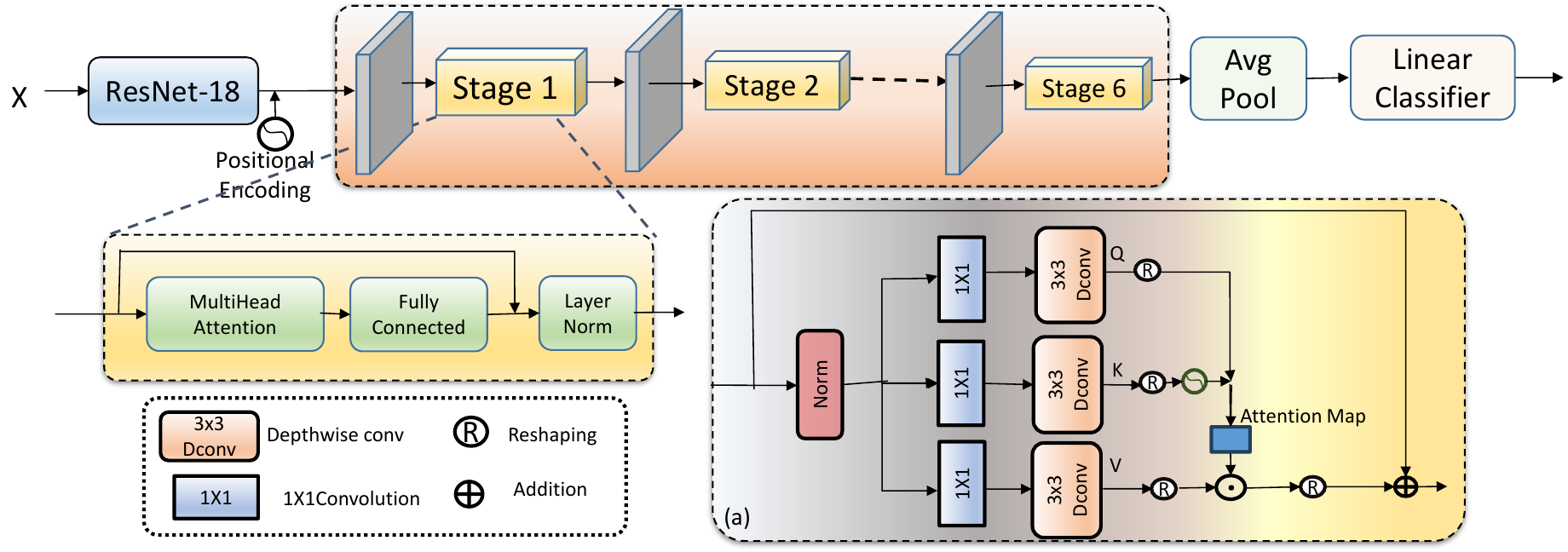}}
\caption{The proposed model with Unified convolutional transformer. First, the input features are given to the convolution block and then through the MHA. a) shows the flow of input from the C-block to the MHA. }
\label{fig1}
\end{figure*}

\section{Method}
In the proposed UniCon-Former, the input is fed to the ResNet-18 model to get the frame-level features of the input gesture sequence, denoted by $F(X)$.   We pass the input from all the convolutional layers of the ResNet-18 model~\cite{he2016deep} and then through the average pool layer, the output features are obtained. These features are fed to the proposed transformer blocks. The input layer of the model is made to adapt for all the input modalities as proposed in~\cite{molchanov2016online}. The output features from ResNet are of dimensions $N = B\times T \times D$, where $B$ is the batch size, $T$ is the number of frames and $D$ is the dimension of features extracted from ResNet-18 which is the same as the input of the first stage of the transformer block, $d_{model} = 512$. For convenience, we represent $B\times T$ as $L$. So, the features now can be represented as  $L \times D$. 

We add positional embedding as in a traditional transformer and fed the input features $F$ as Query (\textbf{Q}), Key (\textbf{K}), and Value (\textbf{V}) to the proposed UniCon-Former model. The UniCon-Former consists of a C-Block and the Multihead attention which creates a pyramid of attention features at each stage.

\subsection{Convolution Transformer Block}
Before the input features $F$ are given to the Multi-head attention block (MHA), they are passed through the convolution block (C-block). All three vectors Query (\textbf{Q}), Key (\textbf{K}), and Value (\textbf{V}) are passed through the C-block. This helps to combine the advantages of the convolution block
i.e. shared weights, local receptive fields, and spatial sub-sampling) with the advantages of the transformer. The C-block comprises the depthwise-separable convolutional operation on the input features given as, 
\begin{equation}
\label{eqn:4}
\textbf{Q} = Conv(F),
\end{equation}
Here, the $Conv$ is the depthwise separable convolution implemented by: Depth-wise Conv,  BatchNorm2d, Point-wise Conv2d. All the 3 vectors are passed from the C-block to give \textbf{Q}, \textbf{K}, and \textbf{V}. The class token~\cite{jeevan2022resource} is added to the resulting vectors which are then passed through the MHA block. 

The attention from these vectors is calculated through MHA. Then, the attention vector is passed through the sequence of 2 linear layers, let us represent it as $FC$ and we use a dropout of 0.1 before and after the  $FC$ layers. Then, we add a skip connection to the output of the $FC$ and normalize the complete output which can be written as, 
\begin{equation}
\label{eqn:40}
E(x) = Norm(x + FC(MultiHead(\textbf{Q},\textbf{K},\textbf{V}))),
\end{equation}
where $E(x)$ is the complete transformer encoder and $x$ is the input feature to UniCon-Former model. The encoded information is then average pooled over all the frames as, 
\begin{equation}
\label{eqn:5}
H(x) = AvgPool(E(x )),
\end{equation}
where $H(x)$ is the average pooling operation over the $m$ frames. The $H(x)$ output is then passed through a linear classifier to get the output probability distribution.

\subsection{Multi-Modal Late Fusion}\label{9}
% Recently, multi-model methods have been applied in various domains. The key aspect behind this method is to leverage multiple modalities to achieve far better results, compared to the model which processes a single modality.

Multimodal methods have recently gained significant popularity and have been applied in various applications. The key aspect behind this method is to leverage multiple modalities to achieve far better results, compared to the model which processes a single modality. Dynamic hand gesture datasets such as NVGesture and Briareo provide images in different modalities (i.e. RGB, Depth, and infrared images) since they are captured using RGB-G sensors.  Following~\cite{d2020transformer}, we adopt a late fusion technique that combines predictions from each modality independently. To produce the final prediction, we choose the maximum probability from all single-modal inputs across different combinations represented as

\begin{equation}
\label{eqn:6}	
y = \arg\max_j \sum_{i}^n P(\omega_j|x_i),
\end{equation}
where $n$ is the number of modalities over which the results are to be aggregated, and  $P(\omega_j|x_i)$ is the probability distribution of the  $i^{th}$ frames of a given input, which belongs to class $\omega_j$.

\section{Experiments and Discussion}\label{6}

\begin{table}[h!]
\caption{Results for different modalities on NVGesture~\cite{molchanov2016online}  and Briareo~\cite{manganaro2019hand} dataset. \# is the number of input modalities used. Bold are the best results obtained for each set of modalities.}
\centering
\begin{tabular}{c|ccccc|cM{1cm}|cM{1cm}}
\hline
\multirow{3}{*}{\#}&\multicolumn{5}{c|}{Input data}& \multicolumn{4}{c}{Accuracy} \\ \cline{2-10}
&\multirow{2}{*}{Color} &\multirow{2}{*}{ Depth}& \multirow{2}{*}{IR} & \multirow{2}{*}{Normals} &{Optical} & \multicolumn{2}{c|}{NVGesture} & \multicolumn{2}{c}{Briareo}\\  \cline{7-10}
&&&&& flow& Transformer~\cite{d2020transformer} &{Ours} & Transformer~\cite{d2020transformer} & {Ours}\\
\hline

\multirow{5}{*}{1} 
&\checkmark &&&&&    76.50\%&81.67\%& 90.60\%&96.53\%  \\
&&\checkmark &&&&    83.00\%&\textbf{85.27}\%& 92.40\%&97.57\%\\
&&&\checkmark&&&    64.70\%&67.63\%& 95.10\%&\textbf{97.92}\%\\
&&&&\checkmark&&    82.40\%&83.61\%& 95.80\%&97.57\%\\ 
&&&&&\checkmark&   72.00\%&74.58\%&-&96.53\%\\ \hline

\multirow{10}{*}{2}
&\checkmark&\checkmark&&&&  84.60\% &87.55\%&94.10\%&97.92\%\\
&\checkmark&&\checkmark&&&  79.00\%&83.40\%&95.50\%&\textbf{98.26}\%\\
&&\checkmark&\checkmark&&&  81.70\%&84.23\%&95.10\%&97.91\%\\
&\checkmark&&&\checkmark&&  84.60\%&\textbf{85.06}\%&96.50\%&96.88\%\\
&&\checkmark&&\checkmark&&  87.30\%&\textbf{85.06}\%&96.20\%&97.57\%\\
&&&\checkmark&\checkmark&&  83.60\%&82.78\%&97.20\%&97.57\%\\ 
&\checkmark&& &&\checkmark& -&83.40\%&-&97.22\%\\
&&\checkmark&&&\checkmark&  -&84.65\%&-&97.56\%\\ 
&&&\checkmark&& \checkmark& -&77.39\%&-&97.91\%\\
&&&&\checkmark&\checkmark& - &84.44\%&-&96.88\%\\\hline

\multirow{10}{*}{3} 
&\checkmark&\checkmark&\checkmark&&& 85.30\% &87.14\%&95.10\%&97.92\%\\
&\checkmark&\checkmark&&\checkmark&& 86.10\%&86.72\%&95.80\%&97.57\%\\
&\checkmark&&\checkmark&\checkmark&& 85.30\%&\textbf{87.34}\%&96.90\%&96.88\%\\
&&\checkmark&\checkmark&\checkmark&& 87.10\%&85.89\%&97.20\%&97.57\%\\ 

&\checkmark&\checkmark&&&\checkmark& -&85.51\%&-&97.92\%\\ 
&\checkmark&&\checkmark&&\checkmark& -&85.06\%&-&\textbf{98.61}\%\\ 
&\checkmark&&&\checkmark&\checkmark& -&87.14\%&-&96.88\%\\
&&\checkmark&\checkmark&&\checkmark& -&84.85\%&-&98.26\%\\ 
&&&\checkmark&\checkmark&\checkmark& -&85.89\%&-&96.87\%\\ 
&&\checkmark&&\checkmark&\checkmark& -&85.89\%&-&97.57\%\\ 
\hline

\multirow{5}{*}{4}
&\checkmark&\checkmark&\checkmark&\checkmark&&  87.60\%&87.14\%&96.20\%&97.57\%\\
&\checkmark&\checkmark&\checkmark&&\checkmark&  -&\textbf{87.97}\%&-&\textbf{98.61}\%\\
&\checkmark&\checkmark&&\checkmark&\checkmark&  -&\textbf{87.97}\%&-&97.92\%\\ 
&\checkmark&&\checkmark&\checkmark&\checkmark&  -&86.72\%&-&96.88\%\\ 
&&\checkmark&\checkmark&\checkmark&\checkmark&  -&87.34\%&-&97.57\%\\
\hline

5&\checkmark&\checkmark &\checkmark&\checkmark&\checkmark& -&\textbf{87.97}\%&-&96.88\%\\
\hline

\end{tabular}
\label{tab1}
\end{table}

\subsection{Implementation Details}\label{7}
% We follow the same training procedure as in~\cite{d2020transformer}, where the model is trained separately for each modality and later fused using late fusion.  
We implemented our work in PyTorch. The model is trained on Nvidia GeForce GTX 1080 Ti GPU hardware. It is optimized using the Adam optimizer with a learning rate of 1e-4 and a weight decay at the 50th and 75th epoch over the categorical cross-entropy loss. Following the approach in~\cite{d2020transformer}, we cropped the image to a size of 224 × 224 pixels to extract features from a pretrained model ResNet-18. To mitigate the over-fitting issue, we used data augmentation techniques such as scaling, cropping, and rotation.  We follow the same training procedure as in~\cite{d2020transformer}, and trained separate models for each modality. Finally, we have applied decision-level fusion using a late fusion approach.

\begin{table}[p]
\caption{Comparison results for single modality on NVGesture dataset~\cite{molchanov2016online}.
% * indicates the model is pre-trained on Kinetics~\cite{kay2017kinetics}, in addition to ImageNet~\cite{5206848}.
}
\centering
\begin{tabular}{ccc}
\hline
Input modality&Method& Accuracy \\ 
\hline
\multirow{15}{*}{Color} 
&Spat. st. CNN~\cite{simonyan2014two} &54.60\%  \\
&iDT-HOG~\cite{wang2016robust}& 59.10\%\\
&C3D~\cite{tran2015learning}&69.30\%\\ 

&R3D-CNN~\cite{molchanov2016online}& 74.10\%\\
&GestFormer~\cite{Garg_2024_CVPR}&75.41\%\\ 
&Res3ATN~\cite{dhingra2019res3atn}& 62.70\%\\
&ConvMixFormer~\cite{garg2024convmixformer}& 76.04\%\\
&PreRNN~\cite{yang2018making}&76.50\% \\
&Transformer~\cite{d2020transformer}& 76.50\%\\
&MVTN~\cite{garg2024mvtnmultiscalevideotransformer}& 77.50\%\\
&I3D~\cite{wang2016robust}&78.40\%\\
&ResNeXt-101~\cite{kopuklu2019real}& 78.63\%\\
&MTUT~\cite{abavisani2019improving}*&81.33\%\\
% &NAS1~\cite{yu2021searching}*&83.61\% \\	
&MsMHA-VTN~\cite{garg2023multiscaled}& 81.42\%\\
&\textbf{UniCon-Former}& \textbf{81.67\%}\\

\hline
&Human~\cite{molchanov2016online}&88.40\% \\  \hline

\multirow{13}{*}{Depth}
& SNV~\cite{yang2014super}& 70.70\%\\
&C3D~\cite{tran2015learning}&78.80\%\\ 
&GestFormer~\cite{Garg_2024_CVPR}&80.21\%\\ 
&R3D-CNN~\cite{molchanov2016online}& 80.30\%\\
&ConvMixFormer~\cite{garg2024convmixformer}& 80.83\%\\
&I3D~\cite{wang2016robust}&82.30\%\\
&Transformer~\cite{d2020transformer}&83.00\%\\
&ResNeXt-101~\cite{kopuklu2019real}& 83.82\%\\
&PreRNN~\cite{yang2018making}&84.40\% \\
&MTUT~\cite{abavisani2019improving}*&84.85\%\\
&MsMHA-VTN~\cite{garg2023multiscaled}& 85.00\%\\
&MVTN~\cite{garg2024mvtnmultiscalevideotransformer}& 85.21\%\\
&\textbf{UniCon-Former}& \textbf{85.27\%}\\
\hline	

\multirow{7}{*}{Optical flow}
&iDT-HOF~\cite{vadisaction}& 61.80\% \\
&Temp. st. CNN~\cite{simonyan2014two} & 68.00\%\\
&Transformer~\cite{d2020transformer}& 72.00\%\\
&MVTN~\cite{garg2024mvtnmultiscalevideotransformer}& 72.50\%\\
&GestFormer~\cite{Garg_2024_CVPR}&72.61\%\\
&ConvMixFormer~\cite{garg2024convmixformer}& 74.17\%\\
% &iDT-MBH~\cite{vadisaction} & 76.80\%\\
% &R3D-CNN~\cite{molchanov2016online} & 77.80\%\\
% &MTUT~\cite{abavisani2019improving}*&83.40\%\\
% &I3D~\cite{wang2016robust} & 83.40\%\\
&\textbf{UniCon-Former}&\textbf{74.58\%}\\
\hline

\multirow{5}{*}{Normals}&ConvMixFormer~\cite{garg2024convmixformer}& 80.21\%\\
&GestFormer~\cite{Garg_2024_CVPR}&81.66\%\\ 
&Transformer~\cite{d2020transformer} &82.40\% \\
&MVTN~\cite{garg2024mvtnmultiscalevideotransformer}& 83.75\%\\
&\textbf{UniCon-Former}& \textbf{83.61\%}\\ \hline

\multirow{6}{*}{Infrared}
&R3D-CNN~\cite{molchanov2016online}& 63.50\%\\
&ConvMixFormer~\cite{garg2024convmixformer}& 63.54\%\\
&GestFormer~\cite{Garg_2024_CVPR}&63.54\%\\ 
& Transformer~\cite{d2020transformer}&  64.70\% \\
&MVTN~\cite{garg2024mvtnmultiscalevideotransformer}& 70.42\%\\
&\textbf{UniCon-Former}&\textbf{67.63\%}\\ \hline
\end{tabular}
\label{tab2}
\end{table}

\begin{table}[h!]
\caption{Comparison results for multi-modal inputs on NVGesture dataset~\cite{molchanov2016online}. *~indicates the model is pre trained on Kinetics~\cite{kay2017kinetics}, in addition to ImageNet~\cite{5206848}.  }
\centering
\begin{tabular}{ccc}
\hline
 Method &Input modality & Accuracy \\ 
\hline
Two-st. CNNs~\cite{simonyan2014two} & color + flow & 65.60\%\\
\hline
iDT ~\cite{vadisaction}& Color + flow & 73.00\% \\
\hline
R3D-CNN~\cite{molchanov2016online} & Color + flow & 79.30\%\\
R3D-CNN~\cite{molchanov2016online} & Color + depth + flow & 81.50\%\\
R3D-CNN~\cite{molchanov2016online} & Color + depth + ir & 82.00\%\\
R3D-CNN~\cite{molchanov2016online} & depth + flow & 82.40\%\\
R3D-CNN~\cite{molchanov2016online} & all & 83.80\%\\
\hline
MSD-2DCNN~\cite{fan2021multi}&Color+depth&84.00\% \\

\hline
8-MFFs-3f1c\cite{kopuklu2018motion}&Color + flow& 84.70\%\\
\hline
STSNN~\cite{zhang2020dynamic}&Color+flow& 85.13\%\\
\hline
PreRNN~\cite{yang2018making}& Color + depth&85.00\% \\
\hline

I3D~\cite{wang2016robust}& Color + depth &83.80\%\\
I3D~\cite{wang2016robust}& Color + flow &84.40\%\\
I3D~\cite{wang2016robust}& Color + depth + flow &85.70\%\\

\hline
GPM~\cite{fan2021multi}& Color + depth&86.10\% \\
\hline
MTUT\textsubscript{RGB-D}~\cite{abavisani2019improving}*& Color + depth& 85.50\%\\
MTUT\textsubscript{RGB-D+flow}~\cite{abavisani2019improving}*& Color + depth& 86.10\%\\
MTUT\textsubscript{RGB-D+flow}~\cite{abavisani2019improving}*& Color + depth + flow& 86.90\%\\
\hline

Transformer~\cite{d2020transformer}& depth + normals &87.30\%\\
Transformer~\cite{d2020transformer}& Color + depth + normals+ir& 87.60\%\\
\hline
NAS2~\cite{yu2021searching}*& Color + depth&86.93\% \\
NAS1+NAS2~\cite{yu2021searching}*&Color + depth&88.38\% \\

\hline
ConvMixFormer&depth + op&82.16\%\\	
ConvMixFormer&depth + normal+ ir\textbf{ }&84.02\%\\ 
{ConvMixFormer}&depth + ir+ normal + flow &{85.49\%}\\ 
\hline

GestFormer~\cite{Garg_2024_CVPR}&depth + normals&82.78\%\\	
GestFormer~\cite{Garg_2024_CVPR}&depth + color + ir\textbf{ }&84.24\%\\ 
{GestFormer~\cite{Garg_2024_CVPR}}&depth + color + ir + normal&{85.62\%}\\ {GestFormer~\cite{Garg_2024_CVPR}}&depth + color + ir + normal + flow&{85.85\%}\\
\hline
MVTN~\cite{garg2024mvtnmultiscalevideotransformer}&depth + normals&85.64\%\\	
MVTN~\cite{garg2024mvtnmultiscalevideotransformer}&depth + color+ ir&87.80\%\\ 
\hline

\textbf{UniCon-Former}&\textbf{Color + normal+ ir }&\textbf{87.34\%}\\ 
\textbf{UniCon-Former}&\textbf{depth + ir + color + flow }&\textbf{87.97\%}\\ 
{\textbf{UniCon-Former}}&{\textbf{depth +  Color + flow + normal}}&{\textbf{87.97\%}}\\ 
{\textbf{UniCon-Former}}&{\textbf{Color + depth + ir  + normal + flow}}&{\textbf{87.97\%}}\\ 

\hline	
\end{tabular}
\label{tab3}
\end{table}

\subsection{Results and Discussion}
\textbf{NVGesture:} 
We compare the performance of the proposed model with the vanilla transformer model used for gesture recognition~\cite{d2020transformer} and the results are shown in Table~\ref{tab1}. For each input modality (color, depth, IR, normals, and optical flow), our model UniCon-Former performs better than the vanilla transformer. From the table, we can see that an increase of 6.76\%, 2.73\%, 4.53\%, 1.47\%, and 3.58\%  is seen when the model is independently trained on color, depth, IR, normals, and optical flow, respectively. This increase is quite considerable, with 85.27\% for the depth modality. 

Further, when we experimented with 2 modalities, we observed incremental accuracy for each experiment with the best accuracy of 85.06\% on depth and normal or color and normal. This is because the model can learn multiscale features over each transformer stage. On further increasing the input modalities, the accuracy increases to 87.34\% with the combined input of color, it, and normal. A little increase in accuracy us observed at 87.98\% for 4 modality inputs which remains the same on further increasing the input modalities. Thus, overall we also observe that the proposed UniCon-Former performs better than the traditional transformer. 

Also, in Table~\ref{tab2} and Table~\ref{tab3}, we compare the proposed model with other methods on single and multimodal inputs, respectively. From table~\ref{tab2}, we can conclude that our model has outperformed other methods expect MVTN~\cite{garg2024mvtnmultiscalevideotransformer} for normals and Infrared.  Similarly, we compare the results for multimodal inputs in Table~\ref{tab3} and conclude that the best accuracy is obtained  with 4 modalities. 

\textbf{Briareo:} We have also experimented with the Briareo dataset and compared the results with traditional transformer~\cite{d2020transformer} as reported in Table~\ref{tab1}. We observe that we have outperformed~\cite{d2020transformer} with 6.55\%, 5.60\%, 2.97\%, 1.85\% increment on color, depth, IR, and normals input, respectively.  We also observe that best accuracy of 97.92\% for single modality is obtained on infrared input and it further increases to 98.26\% with double modality and 98.61\% with 3 modality input.  Also, a comparison with the other methods are shown in Table~\ref{tab5} and it significantly outperforms other methods.

\begin{table}[t]
\caption{Comparison of the results obtained for different modalities on Briareo dataset~\cite{manganaro2019hand}.}
\centering
\begin{tabular}{ccc}
\hline
Method& Input Modality& Accuracy \\ 
\hline
C3D-HG~\cite{manganaro2019hand}& Color& 72.20\%\\
C3D-HG~\cite{manganaro2019hand}& depth& 76.00\%\\
C3D-HG~\cite{manganaro2019hand}& ir& 87.50\%\\
NUI-CNN~\cite{d2020multimodal}& depth & 90.30\%\\
NUI-CNN~\cite{d2020multimodal}& Color& 83.30\%\\
NUI-CNN~\cite{d2020multimodal}& ir& 86.10\%\\
LSTM-HG~\cite{manganaro2019hand}&3D joint features &94.40\%\\
\hline
NUI-CNN~\cite{d2020multimodal}& depth + ir& 92.00\%\\
NUI-CNN~\cite{d2020multimodal}& Color + depth + ir& 90.90\%\\

\hline
Transformer~\cite{d2020transformer}& normals& 95.80\%\\
Transformer~\cite{d2020transformer}& depth + normals &96.20\%\\
Transformer~\cite{d2020transformer}&ir + normals &97.20\%\\
\hline	

GestFormer~\cite{Garg_2024_CVPR} & ir&98.13\%\\
GestFormer~\cite{Garg_2024_CVPR} & ir + normals &97.57\%\\
\hline

MVTN~\cite{garg2024mvtnmultiscalevideotransformer} & normals&98.26 \%\\
MVTN~\cite{garg2024mvtnmultiscalevideotransformer} & color + depth + normals &98.61\%\\
MVTN~\cite{garg2024mvtnmultiscalevideotransformer} &depth + ir + normal &98.61\%\\
\hline

\textbf{UniCon-Former} & \textbf{ir}&\textbf{97.92} \%\\
\textbf{UniCon-Former} & \textbf{Color + ir} &\textbf{98.26}\%\\
\textbf{UniCon-Former} &\textbf{Color+ ir + flow} &\textbf{98.61}\%\\
\textbf{UniCon-Former} &\textbf{Color+ ir + flow + depth} &\textbf{98.61}\%\\
\hline
\end{tabular}
\label{tab5}
\end{table}
\subsection{Parameter Efficiency}
Table~\ref{tab24} compares UniCon-Former with state-of-the-art methods based on the number of parameters and MACs. A trade-off between the complexity and performance concludes that with lesser number of parameters and less complexity our model achieves state-of-the-art results. 

\begin{table}[t]
\caption{Comparison in terms of the number of parameters (M) and MACs. The numbers of MACs are counted by fvcore library. }
\vspace{2.5mm}
\centering
\begin{tabular}{c|cc}
\hline
Methods&Params (M)& MACs (G)\\  \hline
NAS1~\cite{yu2021searching} &93.90&60.44 \\
NAS2~\cite{yu2021searching} &251.40& 116.20\\
ResNeXt-101~\cite{kopuklu2019real} &52.28&- \\
R3D-CNN~\cite{molchanov2016online} & 38.00&-\\
NUI-CNN~\cite{d2020multimodal}& 28.00& -\\
C3D-HG~\cite{manganaro2019hand} & 26.70&-\\
Transformer\cite{d2020transformer} & 24.30&62.92\\
GestFormer~\cite{Garg_2024_CVPR}&24.08&60.40\\
\textbf{UniCon-Former}&\textbf{19.58}&\textbf{60.25}\\
\hline
\end{tabular}
\label{tab24}
\end{table}	

\section{Conclusion}
We proposed a novel Unified Convolution Transformer Network (UniCon-Former)  for dynamic hand gesture recognition that learns the multiscale feature at different stages of the transformer. This helps to tackle the problem of hand shape and size variation by extracting contextual information at different levels in a hierarchical manner which helps to reduce the computational cost. Extensive experiments on the Briareo and NVGesture datasets are performed which shows that our model is better than the traditional transformer model. These results validate the design considerations and highlight the effectiveness of the proposed model for dynamic hand gesture recognition tasks with less model complexity and parameters.

\bibliographystyle{splncs04}
\bibliography{main}
\end{document}